%% file: iccv_co_matching.tex
\documentclass[10pt,twocolumn,letterpaper]{article}

\usepackage{iccv}
\usepackage{times}
\usepackage{epsfig}
\usepackage{graphicx}
\usepackage{amsmath,bm}
\usepackage{amssymb}
\usepackage{array}

\makeatletter  
\newif\if@restonecol  
\makeatother

\usepackage[linesnumbered,ruled,vlined]{algorithm2e}
\usepackage{algpseudocode}  
\usepackage{amsmath}  

\iccvfinalcopy
\usepackage[pagebackref=true,breaklinks=true,letterpaper=true,colorlinks,bookmarks=false]{hyperref}

\def\iccvPaperID{7962} 

\ificcvfinal\fi

\begin{document}

\title{Co-matching: Combating Noisy Labels by Augmentation Anchoring}

\author{Yangdi Lu\\
Mcmaster University\\
{\tt\small luy100@mcmaster.ca}
\and
Yang Bo\\
Mcmaster University\\
{\tt\small boy2@mcmaster.ca}
\and
Wenbo He\\
Mcmaster University\\
{\tt\small hew11@mcmaster.ca}
}

\maketitle
\ificcvfinal\thispagestyle{empty}\fi

\begin{abstract}
   Deep learning with noisy labels is challenging as deep neural networks have the high capacity to memorize the noisy labels. In this paper, we propose a learning algorithm called Co-matching, which balances the consistency and divergence between two networks by augmentation anchoring. Specifically, we have one network generate anchoring label from its prediction on a weakly-augmented image. Meanwhile, we force its peer network, taking the strongly-augmented version of the same image as input, to generate prediction close to the anchoring label. We then update two networks simultaneously by selecting small-loss instances to minimize both unsupervised matching loss (i.e., measure the consistency of the two networks) and supervised classification loss (i.e. measure the classification performance). Besides, the unsupervised matching loss makes our method not heavily rely on noisy labels, which prevents memorization of noisy labels. Experiments on three benchmark datasets demonstrate that Co-matching achieves results comparable to the state-of-the-art methods.
   
\end{abstract}

\input{intro}

\input{relate}

\input{method}

\input{exp}

\section{Conclusion}
In this paper, we introduce Co-matching for combating noisy labels. Our method trains two networks simultaneously on ``small-loss" samples and achieves robustness to noisy labels through augmentation anchoring between two networks. By balancing the divergence and consistency of two networks, Co-matching obtains state-of-the-art performance on benchmark datasets with simulated and real-world label noise. For future work, we are interested in exploring the effectiveness of Co-matching in other domains. 

{\small
\bibliographystyle{ieee_fullname}
\bibliography{egbib}
}

\newpage

\section{Supplementary Material}

\subsection{Datasets}
\begin{figure}[h]
	\begin{center}
		\includegraphics[width=0.8\linewidth]{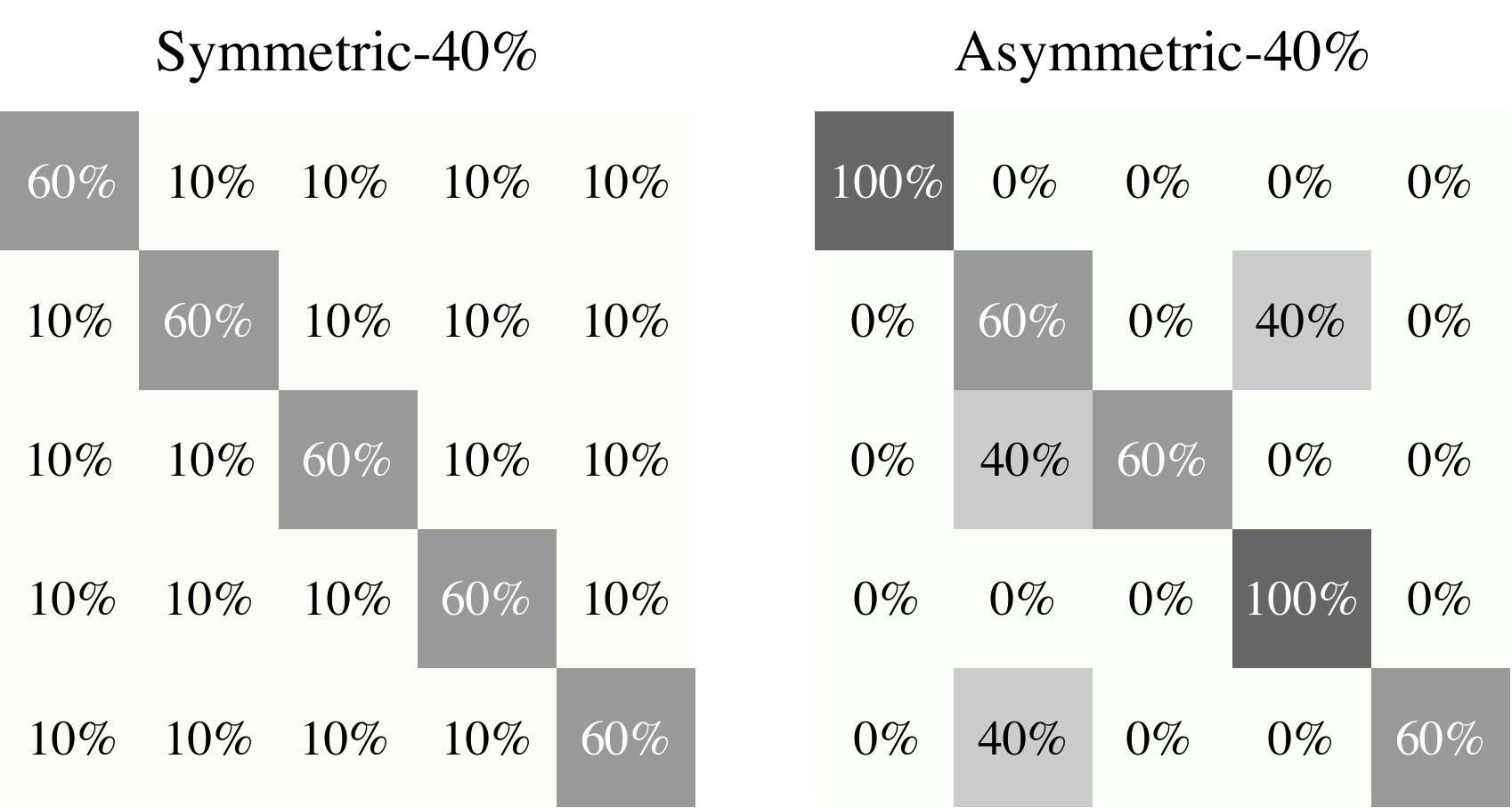}
	\end{center}
	\caption{Example of noise transition matrix $Q$ (taking 5 classes and noise ratio 0.4 as an example).}
	\label{fig:noise_transition}
\end{figure}

The information of datasets are described in Table \ref{tab:dataset}. For Clothing1M, note that we only use 14k and 10k clean data for validation and test. The 50k clean training data is not required during the training. As for simulating label noise, Figure \ref{fig:noise_transition} shows an example of noise transition matrix $Q$. Specifically, for CIFAR-10, the asymmetric noisy labels are generated by flipping \emph{truck} $\rightarrow$ \emph{automobile}, \emph{bird} $\rightarrow$ \emph{airplane}, \emph{deer} $\rightarrow$ \emph{horse} and \emph{cat} $\leftrightarrow$ \emph{dog}. For CIFAR-100, the noise flips each class into the next, circularly within super-classes.

\begin{table}[h]
	\begin{center}
		\resizebox{.45\textwidth}{!}{
			\begin{tabular}{c|c|c|c|c}
				\hline\hline
				& \# of train & \# of test & \# of class & input size  \\
				\hline
				CIFAR-10 & 50,000 & 10,000 & 10 & 32 $\times$ 32 \\
				CIFAR-100 & 50,000 & 10,000 & 100 & 32 $\times$ 32 \\
				Clothing1M & 1,000,000 &10,000&14&224 $\times$ 224\\
				\hline \hline
			\end{tabular}
		}
	\end{center}
	\caption{Summary of datasets used in the experiments.} \label{tab:dataset}
\end{table}

\subsection{Network Architecture}
The network architectures of the CNN model for CIFAR-10 and CIFAR-100 are shown in Table \ref{table:network}.

\begin{table}[h]
	\begin{center}
		\resizebox{0.3\textwidth}{!}{
			\begin{tabular}{c}
				\hline\hline
				CNN on CIFAR-10 \& CIFAR-100 \\
				\hline
				32 $\times$ 32 RGB Image  \\
				\hline
				3 $\times$ 3, 64 BN, ReLU  \\
				3 $\times$ 3, 64 BN, ReLU  \\
				2 $\times$ 2 Max-pool \\
				\hline
				3 $\times$ 3, 128 BN, ReLU  \\
				3 $\times$ 3, 128 BN, ReLU  \\
				2 $\times$ 2 Max-pool \\
				\hline
				3 $\times$ 3, 196 BN, ReLU  \\
				3 $\times$ 3, 196 BN, ReLU  \\
				2 $\times$ 2 Max-pool \\
				\hline
				Dense 256 $\rightarrow$ 100 \\
				\hline \hline
			\end{tabular}
		}
	\end{center}
	\caption{The CNN model used on CIFAR-10 and CIFAR-100} \label{table:network}
\end{table}

\subsection{Hyperparameter Sensitivity}

\noindent\textbf{Choice of $\lambda$.} To conduct the sensitivity analysis on hyperparameter $\lambda$, we set up the experiments on CIFAR-10 and CIFAR-100 with the hardest Symmetric-80\% label noise. Specifically, we compare the $\lambda$ in the range of $[0.05,0.35,0.65,0.95]$. Recall that $\lambda$ controls the importance weights of classification loss and matching loss. A larger $\lambda$ means a larger contribution from the matching loss.

\begin{figure}[h]
	\begin{center}
		\includegraphics[width=0.99\linewidth]{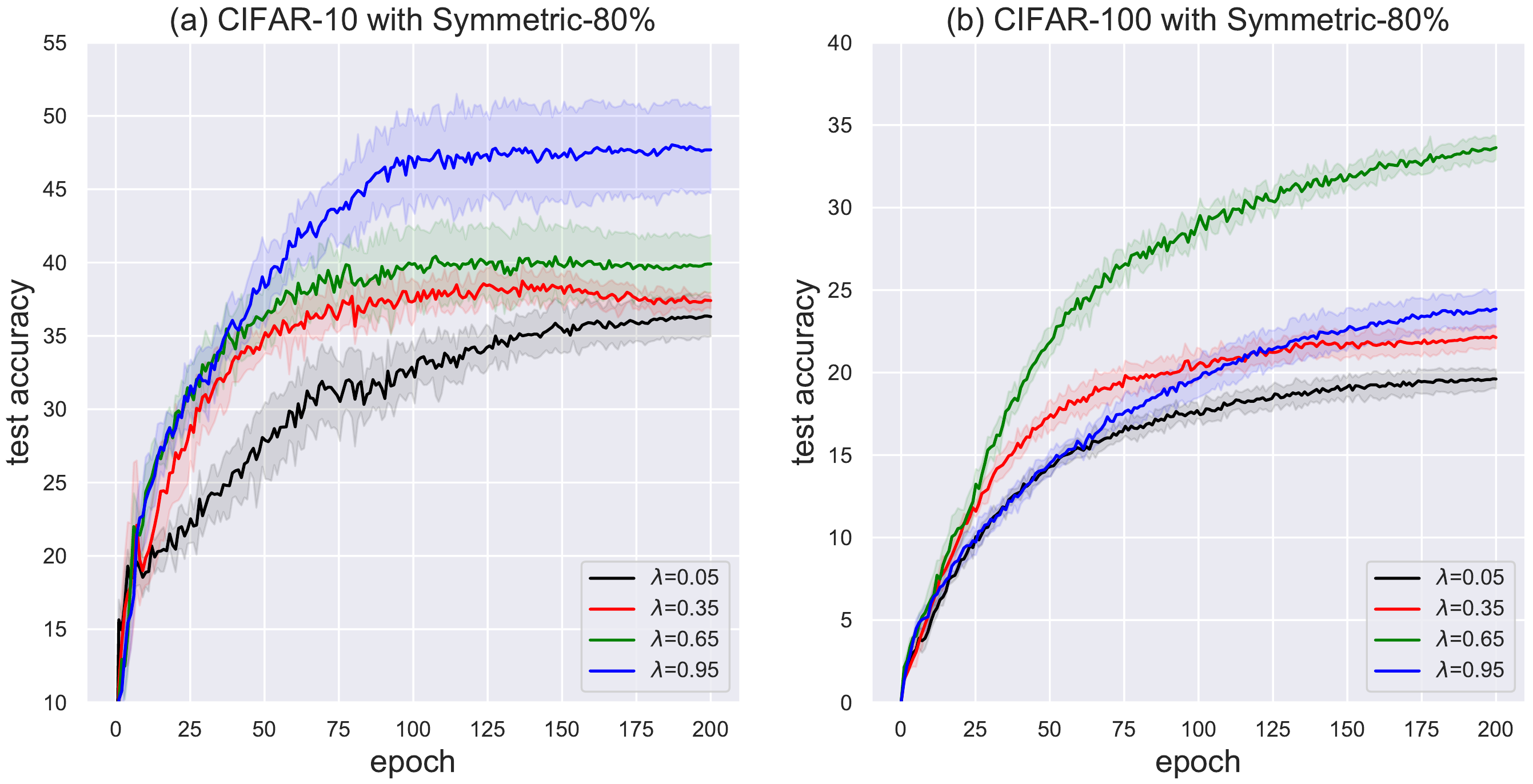}
	\end{center}
	\caption{Test accuracy of Co-matching with different $\lambda$ on CIFAR-10 and CIFAR-100. $\tau=0.6\epsilon$ for both cases.} 
	\label{fig:lambda}
\end{figure}

Figure \ref{fig:lambda} shows the test accuracy vs. number of epochs. In the CIFAR-10 Symmetric-80\% case, a larger $\lambda$ gets a better test accuracy, and $\lambda=0.95$ achieves the best performance. It verifies the motivation of the loss function in Co-matching: under high-levels of label noise, the model is hard to get enough supervision from noisy labels if we only use the classification loss, more weights on unsupervised loss is required to improve the generalization ability. $\lambda=0.65$ is the best option in the CIFAR-100 Symmetric-80\% case. Because different datasets require different significance of augmentation anchoring, too much weights on unsupervised loss may cause the model hard to converge.


\begin{figure}[t]
	\begin{center}
		\includegraphics[width=0.99\linewidth]{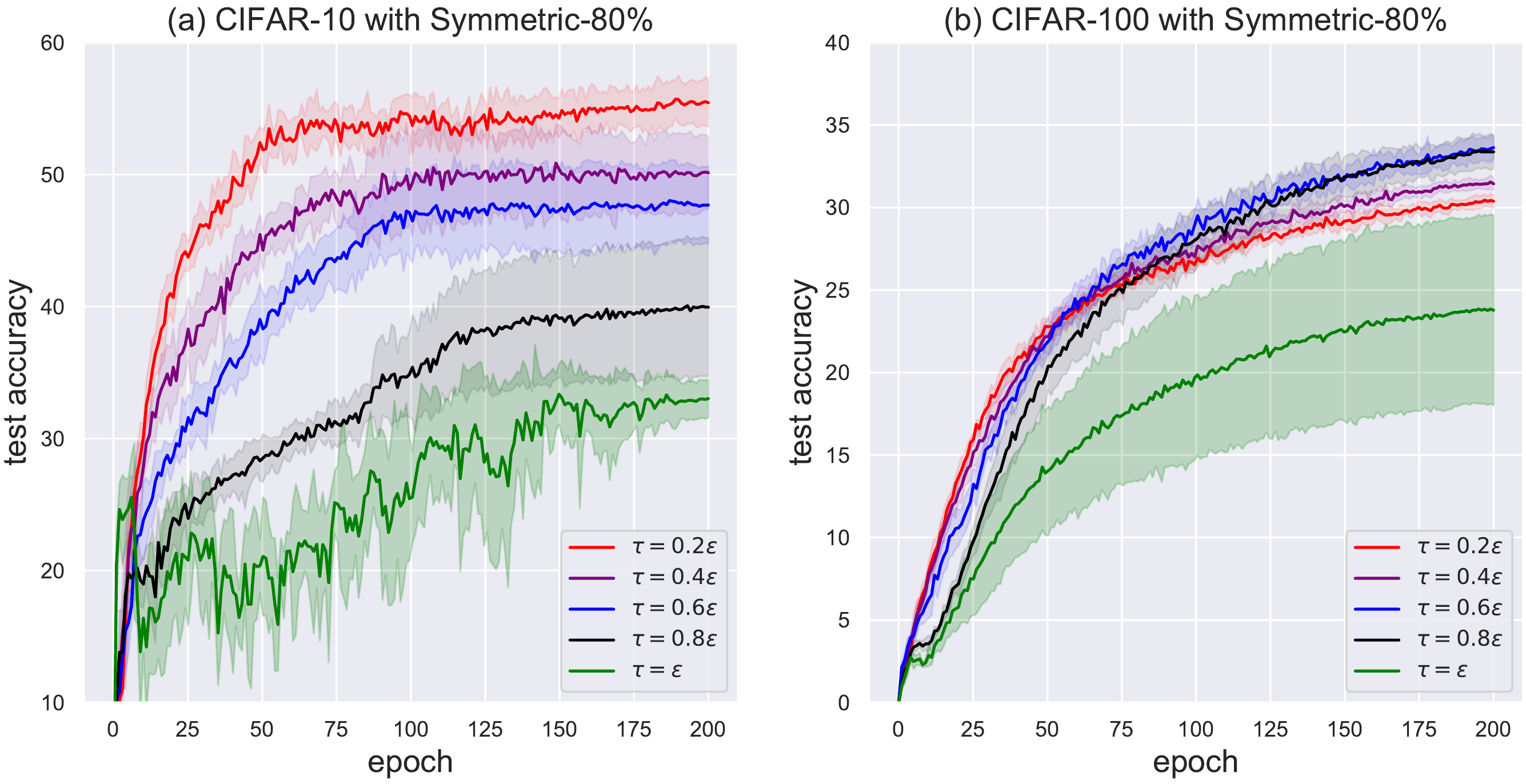}
	\end{center}
	\caption{Test accuracy of Co-matching with different $\tau$ on CIFAR-10 and CIFAR-100. $\lambda=0.95$ for plot (a) and $\lambda=0.65$ for plot (b).} 
	\label{fig:epsilon}
\end{figure}

\noindent\textbf{Choice of $\tau$.} The experiments are also conducted on CIFAR-10 and CIFAR-100 with the hardest Symmetric-80\% case. Specifically, we compare the $\tau$ in the range of $[0.2\epsilon,0.4\epsilon,0.6\epsilon,0.8\epsilon,\epsilon]$, where $\epsilon$ is the inferred noise rate that widely used in previous works as we have discussed in the paper. Recall that smaller $\tau$ in $R(t)$ increases the number of small-loss instances to be selected. As shown in Figure \ref{fig:epsilon}, in CIFAR-10 with Symmetric-80\% case, the smaller $\tau$ achieves better and better performance. In CIFAR-100 with Symmetric-80\%, $\tau=0.6\epsilon$ is the best option. Overall, under a high-level of label noise, since the augmentation anchoring component requires enough samples for effective training, a proper increase of the selected small-loss instances achieves better generalization ability.






\subsection{State-of-the-art methods with the same weak and strong augmentation}

\begin{figure}[h]
	\begin{center}
		\includegraphics[width=0.99\linewidth]{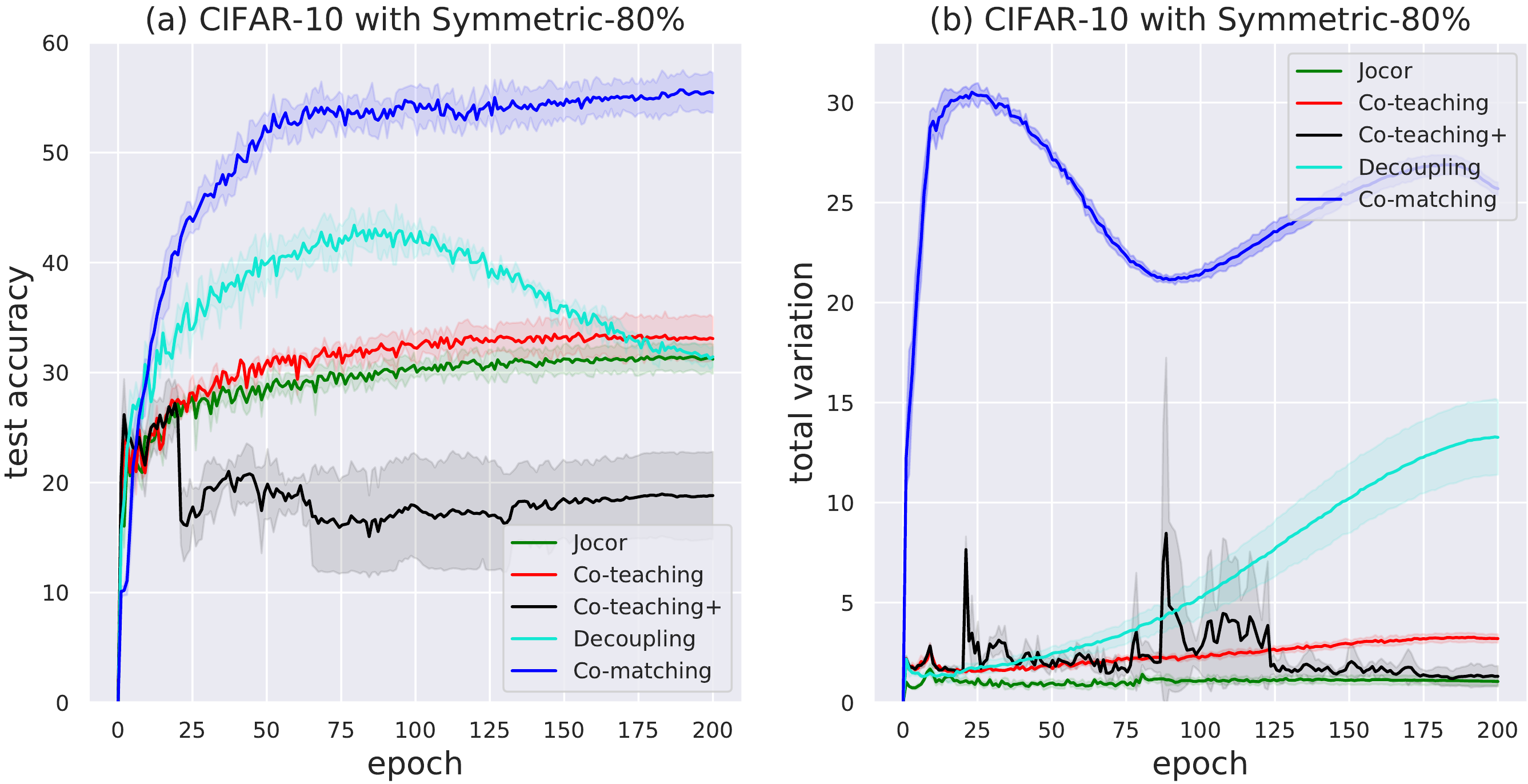}
	\end{center}
	\caption{(a) Test accuracy of state-of-the-art methods with same weak and strong augmentation on CIFAR-10. (b) Total variation of state-of-the-art methods with same weak and strong augmentation on CIFAR-10. } 
	\label{fig:sam_aug}
\end{figure}

We evaluate the state-of-art methods with the same augmentation strategy (i.e. weak and strong for each network respectively) as Co-matching. We set up the experiments on CIFAR-10 with the hardest Symmetric-80\% label noise. The weak augmentation is a standard crop-and-flip. The strong augmentation is crop-and-flip and RandAugment with $M=2$. Figure \ref{fig:sam_aug} (a) shows the results. We observe that with the same augmentation strategy, the performance of state-of-the-art methods do not improve a lot. Co-teaching+ even performs worse. Besides, the state-of-the-art methods become more stable as we can see the standard deviation from the mean becomes smaller.

\subsection{Divergence of two networks}

We compare the divergence of two networks by reporting total variations of softmax output between two networks in Figure \ref{fig:sam_aug} (b). We can clearly observe that two networks trained by Co-teaching and Jocor converge to a consensus gradually, while the two networks trained by Co-matching keeps diverged.

\subsection{Using hard pseudo-labeling for matching loss}

Using hard pseudo-labeling for matching loss helps the model converge. We report the results in Table \ref{tab:hard}. We find that when the noise reaches to 80\%, the Co-matching does not converge with soft pseudo-labeling. That's the main reason we use hard pseudo-labeling for matching loss.

\begin{table}[h]
	\begin{center}
		\resizebox{.45\textwidth}{!}{
			\begin{tabular}{c|c|c}
				\hline\hline
				& Hard pseudo-labeling & Soft pseudo-labeling   \\
				\hline
				Symmetric-50\% & 86.42 $\pm$ 0.18 & 86.43 $\pm$ 0.32  \\
				Symmetric-80\% & 55.42 $\pm$ 1.68 & 9.98 $\pm$ 0.0  \\
				\hline \hline
			\end{tabular}
		}
	\end{center}
	\caption{Average test accuracy (\%) on CIFAR-10.  } \label{tab:hard}
\end{table}

\subsection{List of Data Augmentations}

We use the same sets of image transformation used in RandAugment. For completeness, we describe the all transformation operations for these augmentation strategies in Table \ref{table:aug_list}.

\begin{table*}[h]
	\begin{center}
		\resizebox{0.9\textwidth}{!}{
			\begin{tabular}{lp{10cm}cc}
				\hline\hline
				Transformations & Description& Parameter&Range \\
				\hline
				Autocontrast & Maximizes the image contrast by setting the darkest (lightest) pixel to black (white). & & \\
				\hline
				Brightness & Adjusts the brightness of the image. $B = 0$ returns a black image, $B = 1$ returns the original image.  & $B$& $[0.05,0.95]$\\
				\hline
				Color &  Adjusts the color balance of the image like in a TV. $C=0$ returns a black\& white image, $C=1$ returns the original image. & $C$& $[0.05,0.95]$\\
				\hline
				Contrast & Controls the contrast of the image. $C = 0$ returns a gray image, $C = 1$ returns the original image.  & $C$&  $[0.05,0.95]$\\
				\hline
				Sharpness & Adjusts the sharpness of the image, where $S = 0$ returns a blurred image, and $S = 1$ returns the original image.  & $S$&$[0.05,0.95]$\\ 
				\hline
				Gaussian Blur & Blur the image by Gaussian function && \\
				\hline
				Solarize & Inverts all pixels above a threshold value of $T$. &$T$& $[0,1]$\\ 
				\hline
				Posterize & Reduces each pixel to $B$ bits.  & $B$& $[4,8]$\\
				\hline
				Equalize & Equalizes the image histogram.  &&\\
				\hline
				Identity & Returns the original image. & & \\	
				\hline
				Rotate &Rotates the image by $\theta$ degrees.  & $\theta$& $[-30,30]$\\	
				\hline
				ShearX & Shears the image along the horizontal axis with rate $R$.  & $R$& $[-0.3,0.3]$\\
				\hline
				ShearY & Shears the image along the vertical axis with rate $R$.  & $R$& $[-0.3,0.3]$\\	
				\hline
				TranslateX & Translates the image horizontally by ($\lambda \times$ image width) pixels.& $\lambda$& $[-0.3,0.3]$  \\
				\hline
				TranslateY & Translates the image vertically by ($\lambda \times$ image height) pixels.& $\lambda$& $[-0.3,0.3]$  \\
				\hline \hline
			\end{tabular}
		}
	\end{center}
	\vspace{-0.5em}
	\caption{List of transformations used in RandAugment.} \label{table:aug_list}
\end{table*}

\end{document}

%% file: intro.tex
\section{Introduction}

Deep Neural Networks (DNNs) have shown remarkable performance in a variety of applications \cite{krizhevsky2012imagenet,liu2019improving,szegedy2015going}. However, the superior performance comes with the cost of requiring a correctly annotated dataset, which is extremely time-consuming and expensive to obtain in most real-world scenarios. Alternatively, we may obtain the training data with annotations efficiently and inexpensively through either online key search engine \cite{li2017webvision} or crowdsourcing \cite{yu2018learning}, but noisy labels are likely to be introduced consequently. Previous studies \cite{arpit2017closer,zhang2016understanding} demonstrate that fully memorizing noisy labels affects accuracy of DNNs significantly, hence it is desirable to develop effective algorithms for learning with noisy labels.

To handle noisy labels, most approaches focus on estimating the noise transition matrix \cite{goldberger2016training,patrini2017making,sukhbaatar2014learning,xia2019anchor} or correcting the label according to model prediction \cite{ma2018dimensionality,reed2014training,tanaka2018joint,yi2019probabilistic}. However, it is challenging to estimate the noise transition matrix especially when the number of classes is large. Another promising direction of study proposes to train two networks on small-loss instances \cite{chen2019understanding,han2018co,wei2020combating,yu2019does}, wherein Decoupling \cite{malach2017decoupling} and Co-teaching+ \cite{yu2019does} introduce the ``Disagreement" strategy to keep the two networks diverged to achieve better ensemble effects. However, the instances selected by ``Disagreement" strategy are not guaranteed to have correct labels \cite{han2018co,wei2020combating}, resulting in only a small portion of clean instances being utilized in the training process. Co-teaching \cite{han2018co} and JoCoR \cite{wei2020combating} aim to reduce the divergence between two different networks so that the number of clean labels utilized in each mini-batch increases. In the beginning, two networks with different learning abilities filter out different types of error. However, with the increase of training epochs, two networks gradually converge to a consensus and even make the wrong predictions consistently. 

To address the above concerns, it is essential to keep a balance between divergence and consistency of the two networks. Inspired by augmentation anchoring \cite{berthelot2019remixmatch,sohn2020fixmatch} from semi-supervised learning, we propose a method using weak (e.g. using only crop-and-flip) and strong (e.g. using RandAugment \cite{cubuk2020randaugment}) augmentations for two networks respectively to address the consensus issue. Specifically, one network produces the anchoring labels based on weakly-augmented images. The anchoring labels are used as targets when the peer network is fed the strongly-augmented version of the same images. Their difference is captured by an unsupervised matching loss. Stronger augmentation results in disparate predictions, which guarantees the divergence between two networks, unless they have learned robust generalization ability. As \emph{early-learning} phenomenon shows that the networks fit training data with clean labels before memorizing the samples with noisy labels \cite{arpit2017closer}. Co-matching trains two networks with a loss calculated by interpolating between two loss terms: 1) A supervised classification loss encourages learning from clean labels during the early-learning phase. 2) An unsupervised matching loss limits the divergence of two networks and prevents memorization of noisy labels after the \emph{early-learning} phase. In each training step, we use the small-loss trick to select the most likely clean samples, thus ensuring the error flow from the biased selection would not be accumulated. 




 
	
To show that Co-matching improves the robustness of deep learning on noisy labels, we conduct extensive experiments on both synthetic and real-world noisy datasets, including CIFAR-10, CIFAR-100 and Clothing1M datasets. Experiments show that Co-matching significantly advances state-of-the-art results with different types and levels of label noise. Besides, we study the impact of data augmentation and provide ablation study to examine the effect of different components in Co-matching.

%







%% file: relate.tex
\section{Related work}
\label{sec:relate}
\subsection{Learning with Noisy labels}

\noindent \textbf{Curriculum learning.} Inspired from human cognition, Curriculum learning (CL) \cite{bengio2009curriculum} proposes to start from easy samples and go through harder samples to improve convergence and generalization. In the noisy label scenario, easy (hard) concepts are associated with clean (noisy) samples. Based on CL, \cite{ren2018learning} leverages an additional validation set to adaptively assign weights to noisy samples for less loss contribution in every iteration. 

\noindent\textbf{Small-loss selection.} Another set of emerging methods aim to select the clean labels out of the noisy ones to guide the training. Previous work \cite{arpit2017closer} empirically demonstrates the \emph{early-learning} phenomenon that DNNs tend to learn clean labels before memorizing noisy labels during training, which justifies that instances with small-loss values are more likely to be clean instances. Based on this observation, \cite{lyu2019curriculum} proposes a curriculum loss that chooses samples with small-loss values for loss calculation. MentorNet \cite{jiang2018mentornet} pre-trains a mentor network for selecting small-loss instances to guide the training of the student network. Nevertheless, similar to Self-learning approach, MentorNet inherits the same inferiority of accumulated error caused by the sample-selection bias.
\begin{figure}[t]
	\begin{center}
		\includegraphics[width=1.0\linewidth]{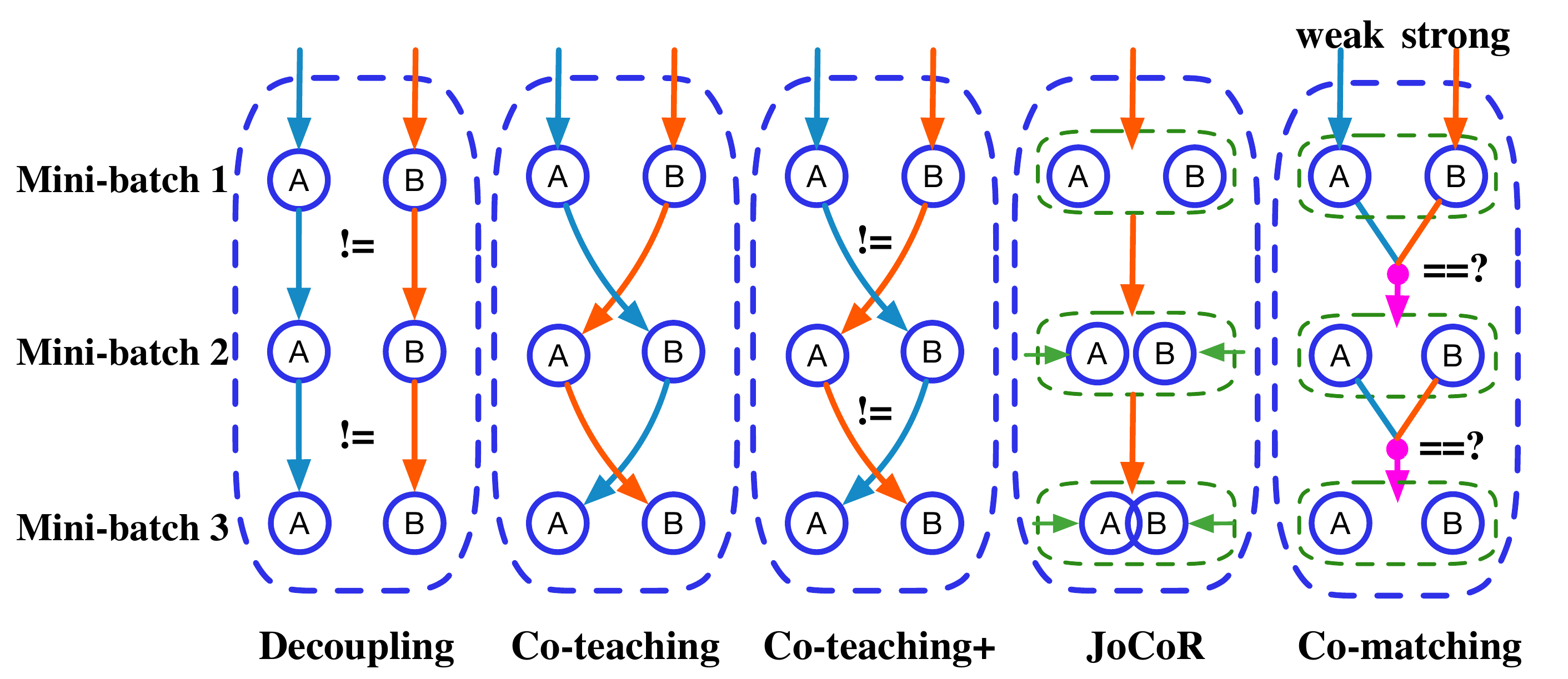}
	\end{center}
	\caption{Comparison of error ﬂow among Decoupling \cite{malach2017decoupling}, Co-teaching \cite{han2018co}, Co-teaching+ \cite{yu2019does}, JoCoR \cite{wei2020combating} and Co-matching. Assume that the error ﬂow comes from the biased selection of training instances, and error ﬂow from network A or B is denoted by blue arrows or red arrows, respectively. \textbf{First panel:} Decoupling maintains two networks (A\&B). The parameters of two networks are updated, when the predictions of them disagree (!=). \textbf{Second panel:} In Co-teaching, each network selects its small-loss data to teach its peer network for the further training. \textbf{Third panel:} In Co-teaching+, each network teaches its small-loss instances with prediction disagreement (!=) to its peer network. \textbf{Fourth panel:} JoCoR trains two networks jointly by using small-loss instances with prediction agreement to make two networks more similar with each other. \textbf{Fifth panel:} On one hand, we keep two networks diverged by feeding images with varying kinds of augmentations. On the other hand, the networks are trained together by minimizing the matching loss to limit their divergence.}
		
	\label{fig:comparison}
\end{figure}

\noindent\textbf{Two classifiers with Disagreement and Agreement.} Inspired by Co-training \cite{blum1998combining}, Co-teaching \cite{han2018co} symmetrically trains two networks by selecting small-loss instances in a mini-batch for updating the parameters. These two networks could filter different types of errors brought by noisy labels since they have different learning abilities. When the error from noisy data flows into the peer network, it will attenuate this error due to its robustness \cite{han2018co}. However, two networks converge to a consensus gradually with the increase of epochs. To tackle this issue, Decoupling \cite{malach2017decoupling} and Co-teaching+ \cite{yu2019does} introduce the ``Update by Disagreement" strategy which conducts updates only on selected instances, where there is a prediction disagreement between two classifiers. Through this, the decision of ``when to update" depends on a disagreement between two networks instead of depending on the noisy labels. As a result, it would reduce the dependency on noisy labels as well as keep two networks divergent. However, as noisy labels are spread across the whole space of examples, there may be very few clean labels in the disagreement area. Thus, JoCoR \cite{wei2020combating} suggests jointly training two networks with the instances that have prediction agreement between two networks. However, the two networks in JoCoR are also prone to converge to a consensus and even make the same wrong predictions when datasets are under high noise ratio. This phenomenon will be explicitly shown in our experiments in the symmetric-80\%. We compare all these approaches in Figure \ref{fig:comparison}.

\noindent\textbf{Other methods.} Some approaches focus on creating noise-tolerant loss functions \cite{ghosh2017robust,van2015learning,wang2019symmetric,zhang2018generalized}. Other methods attempt to adjust the loss \cite{arazo2019unsupervised,hendrycks2018using,ma2018dimensionality,patrini2017making,reed2014training,song2019selfie,tanaka2018joint,wang2018iterative}.
Many approaches \cite{ding2018semi,li2019dividemix,nguyen2020self,tanaka2018joint,wei2020combating} have been proposed to combat noisy labels through semi-supervised learning.

\begin{figure*}[t]
	\begin{center}
		\includegraphics[width=0.8\linewidth]{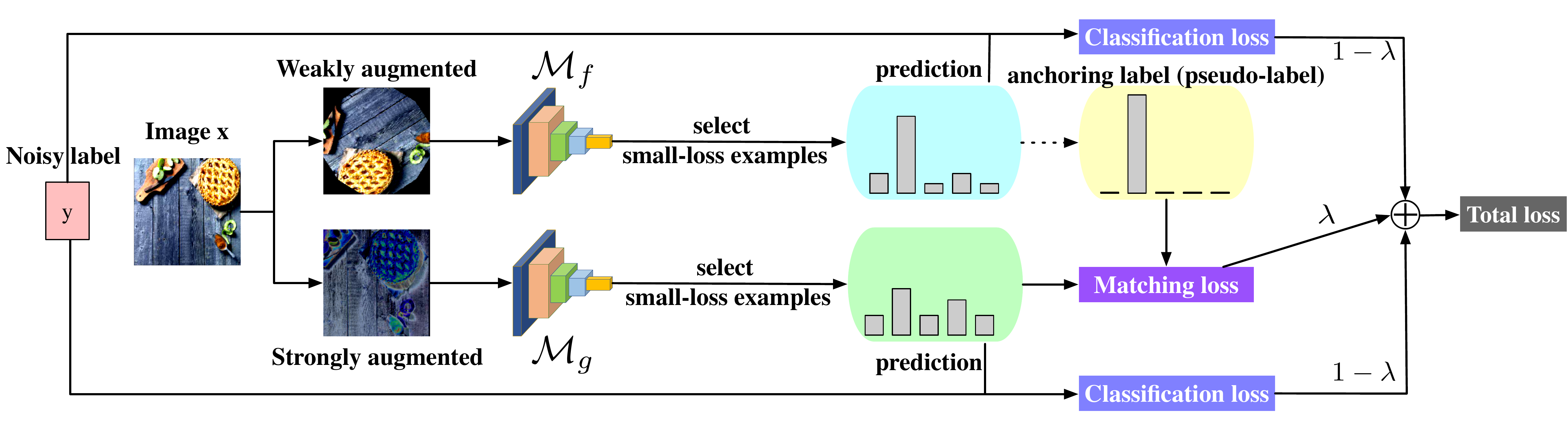}
	\end{center}
	\caption{Diagram of Co-matching. First, a weakly-augmented version of an image $x$ (top) is fed into the model $\mathcal{M}_{f}$ to obtain its prediction (blue box). We convert the prediction to a one-hot pseudo-label as an anchoring label (yellow box). Then, we compute second model's prediction for a strong augmented version of the same image (bottom). The second model is trained to make its prediction on the strongly-augmented version match the anchoring label via an unsupervised matching loss.}
	\label{fig:framework}
\end{figure*}

\subsection{Semi-supervised learning}
Semi-supervised learning provides a means of leveraging unlabeled data to improve a model's performance when only limited labeled data is available. Current methods follow into three classes: \emph{pseudo-labeling} \cite{mclachlan1975iterative,scudder1965probability} leverages the idea that we should use the model itself to obtain artificial labels for unlabeled data; \emph{consistency regularization} \cite{bachman2014learning,laine2016temporal,sajjadi2016regularization} forces the model to produce consistent predictions on perturbed versions of the same image; \emph{entropy minimization} \cite{grandvalet2005semi} encourages the model's output distribution to be low entropy (i.e., make ``high-confidence" predictions) on unlabeled data. Recent works \cite{berthelot2019remixmatch,sohn2020fixmatch} apply augmentation anchoring to replace the consistency regularization as it shows stronger generalization ability.

%% file: method.tex
\section{Method}
\label{sec:method}
In this section, we describe the main components of the proposed approach Co-matching, including structure, loss function, parameter update and augmentation. In addition, we discuss the relations between Co-matching and other existing approaches.	
\subsection{Our algorithm: Co-matching}
\label{sec:comaching}
\noindent \textbf{Notations.} Consider the $C$-class classification problem, we have a training set $D=\{(\bm{x}_{i},\bm{y}_{i})\}^{N}_{i=1}$ with $\bm{y}_{i}\in \{0,1\}^{C}$ being the one-hot vector corresponding to $\bm{x}_{i}$. In the noisy label scenario, $\bm{y}_{i}$ is of relatively high probability to be wrong. We denote the two deep networks in Co-matching as $\mathcal{M}_{f}$ and $\mathcal{M}_{g}$ with parameters $w_{f}$ and $w_{g}$ respectively. $\mathcal{M}_{f}(\bm{x}_{i})$ and $\mathcal{M}_{g}(\bm{x}_{i})$ are the softmax probabilities produced by $\mathcal{M}_{f}$ and $\mathcal{M}_{g}$. For the model inputs, we perform two types of augmentation for each network as part of Co-matching: weak and strong, denoted by $\alpha(\cdot)$ and $\mathcal{A}(\cdot)$ respectively. We will describe the forms of augmentation used for $\mathcal{A}$ and $\alpha$ in section \ref{sec:aug}. 

\noindent \textbf{Structure.} A diagram of Co-matching is shown in Figure \ref{fig:framework}. Co-matching trains two networks simultaneously, as two networks have different abilities to filter out different types of error. By doing this, it suppresses the accumulated error in self-training approach \cite{jiang2018mentornet} with single network. 

\noindent\textbf{Loss function.} \cite{arpit2017closer,zhang2016understanding} experimentally demonstrate that standard cross-entropy loss easily fits the label noise, as its target distribution heavily depends on noisy labels, resulting in undesirable classification performance. To reduce the dependency on noisy labels, the loss function in Co-matching exclusively consists of two loss terms: a supervised loss $\ell_{c}$ for classification task and an unsupervised matching loss $\ell_{a}$ for augmentation anchoring. Our total loss on $\bm{x}_{i}$ is calculated as follows:
\begin{align}
\label{eq1}
	\ell(\bm{x}_{i}) = (1- \lambda)\ell_{c}(\bm{x}_{i},\bm{y}_{i})+\lambda\ell_{a}(\bm{x}_{i}),
\end{align}
where $\lambda\in [0,1]$ is a fixed scalar hyperparameter controlling the importance weight of the two loss terms. Since there are correctly labeled information remaining in noisy labels, classification loss $\ell_{c}$ is the standard cross-entropy loss on noisy labeled instances. 
 \begin{align}
 \label{eq2}
	\ell_{c}(\bm{x}_{i},\bm{y}_{i}) & =  \ell_{\mathcal{M}_{f}}(\bm{x}_{i},\bm{y}_{i})+\ell_{\mathcal{M}_{g}}(\bm{x}_{i},\bm{y}_{i}) \nonumber\\
	& = - \sum_{i=1}^{N}\bm{y}_{i}\log(\mathcal{M}_{f}(\alpha(\bm{x}_{i}))) \\
	&  - \sum_{i=1}^{N}\bm{y}_{i}\log(\mathcal{M}_{g}(\mathcal{A}(\bm{x}_{i}))). \nonumber
\end{align}

For an image $\bm{x}_{i}$, we feed two networks with different levels of augmented images $\alpha(\bm{x}_{i})$ and $\mathcal{A}(\bm{x}_{i})$ respectively, and stronger augmented image $\mathcal{A}(\bm{x}_{i})$ generates disparate prediction compare to the weak one. In this way, Co-matching can always keep two networks diverged throughout the whole training to achieve better ensemble effects. Nonetheless, solely keeping the divergence of two networks will not promote the learning ability to select clean labels by ``small-loss" trick, which is the main drawback of Co-teaching+ as we discussed in Section \ref{sec:relate}. 

To ensure that Co-matching selects more clean labels, we maximize consistency of the two networks as it helps the model find a much wider minimum and provides better generalization performance. In Co-matching, we compute an \emph{anchoring label} for each sample by the prediction of model $\mathcal{M}_{f}$. To obtain an anchoring label, we first compute the predicted class distribution of model $\mathcal{M}_{f}$  given a weakly-augmented version of the image: $\bm{p}_{i}=\mathcal{M}_{f}(\alpha(\bm{x}_{i}))$ and $\bm{p}_{i}=[p_{i}^{1},p_{i}^{2},\cdots,p_{i}^{C}]$. Then, we use hard pseudo-labeling way to get $\hat{\bm{p}}_{i}$ as the anchoring label.
\begin{align}
	\hat{p}_{i}^{j}=\left\{
	\begin{aligned}
		1  & &\text{if}\ j=\arg\max _{c}p_{i}^{c} \\
		0  & &\text{otherwise}
	\end{aligned}
	\right.
\end{align}
The use of hard pseudo-labeling has the similar function to entropy maximization \cite{grandvalet2005semi}, where the model's predictions are encouraged to be low-entropy (i.e., high-confidence). The anchoring label is used as the target for a strongly-augmented version of same image in $\mathcal{M}_{g}$.  We use the standard cross-entropy loss rather than mean squared error or Jensen-Shannon Divergence as it maintains stability and simplifies implementation. Thus, the unsupervised matching loss is 
\begin{align}
	\ell_{a}(\bm{x}_{i}) = - \sum_{i=1}^{N} \bm{\hat{p}}_{i} \log (\mathcal{M}_{g}(\mathcal{A}(\bm{x}_{i}))).
\end{align}
Due to the gradient of cross-entropy loss is
\begin{align}
\nabla \ell_{a}(\bm{x}_{i}) = \sum_{i=1}^{N} \nabla \mathcal{N}_{g}(\mathcal{A}(\bm{x}_{i}))\Big(\mathcal{M}_{g}(\mathcal{A}(\bm{x}_{i}))-\bm{\hat{p}}_{i}\Big),
\end{align}
where $\nabla\mathcal{N}_{g}(\mathcal{A}(\bm{x}_{i}))$ is the Jacobian matrix of the neural network $C$ dimensional encoding $\mathcal{N}_{g}(\mathcal{A}(\bm{x}_{i}))$ for the $i$th input and $\mathtt{softmax}(\mathcal{N}_{g}(\mathcal{A}(\bm{x}_{i})))=\mathcal{M}_{g}(\mathcal{A}(\bm{x}_{i}))$.  Using hard pseudo-labeling helps the model converge faster as it keeps $\mathcal{M}_{g}(\mathcal{A}(\bm{x}_{i}))-\bm{\hat{p}}_{i}$ large before the \emph{early-learning} stage. By minimizing the total loss in (\ref{eq1}), the model consistently improves the generalization performance under different levels of label noise. Under low-level label noise (i.e., 20\%), supervised loss $\ell_{c}(\bm{x}_{i},\bm{y}_{i})$ takes the lead. Co-matching tends to learn from the clean labels by filtering out noisy labels. Besides, Co-matching gets a better ensemble effect due to the divergent of two networks created by feeding different kinds of augmentations. Under high-level label noise  (i.e., 80\%), unsupervised matching loss  $\ell_{a}(\bm{x}_{i})$ takes the lead such that Co-matching inclines to maximize the consistency of the networks to improves the generalization performance without requiring a large number of clean labels.




\noindent\textbf{Parameter update.} DNNs learn clean and easy patterns before memorizing noisy labels \cite{arpit2017closer}. Thus, small-loss instances are more likely to be the ones that are correctly labeled \cite{han2018co}. If we train our classifier to only use small-loss instances in each mini-batch data, it would be resistant to noisy labels. Following Co-teaching, we introduce $R(t)$ to control the number of small-loss instances selected in each mini-batch. 
Intuitively, at the beginning of training, we keep more small-loss instances (with large $R(t)$) in each mini-batch. As DNNs fit noisy labels gradually, we decrease the $R(t)$ quickly at the first $T_{k}$ epochs to prevent networks from over-fitting to the noisy labels. We then conduct small-loss selection as follows:
\begin{align}
	\label{eq5}
	\hat{D}_{n}=\arg\min{_{D'_{n}:|D'_{n}|\geq R(t)|D_{n}|}}\ell(D'_{n}).
\end{align}

\begin{algorithm} [t]
	\nonumber
	\small
	\caption{Co-matching}
	\label{alg:1}
	\KwIn{two networks $\mathcal{M}_{f}$ and $\mathcal{M}_{g}$ with parameters $w=\{w_{f}, w_{g}\}$, weak augmentation $\alpha(\cdot)$, strong augmentation $\mathcal{A}(\cdot)$, training set $D$, batch size $B$, fixed $\tau$, learning rate $\eta$, epoch $T_{k}$, $T_{max}$;}
	\For{$t=1,2,\dots,T_{max}$}
	{
		\textbf{Shuffle} $D$ into $\frac{|D|}{B}$ mini-batches \;
		\For{$n=1,2,\dots,\frac{|D|}{B}$}
		{
			\textbf{Fetch} $n$-th mini-batch $D_{n}$ from $D$ \;
			\textbf{Calculate} the prediction $\mathcal{M}_{f}(\alpha(\bm{x}))$, $\forall \bm{x} \in D_{n}$ \;
			\textbf{Calculate} the prediction $\mathcal{M}_{g}(\mathcal{A}(\bm{x}))$, $\forall \bm{x} \in D_{n}$ \;
			
			\textbf{Calculate} 
			loss $\ell$ by (\ref{eq1}) using $\mathcal{M}_{f}(\alpha(\bm{x}))$ and $\mathcal{M}_{g}(\mathcal{A}(\bm{x}))$, $\forall \bm{x} \in D_{n}$ \;
			
			\textbf{Obtain} small-loss set $\hat{D}_{n}=\arg\min{_{D'_{n}:|D'_{n}|\geq R(t)|D_{n}|}}\ell(D'_{n})$ \;
			
			\textbf{Calculate} $\mathcal{L}_{n} = \frac{1}{|\hat{D}_{n}|}{\sum}_{{\bm{x} \in \hat{D}_{n}}}\ell(\bm{x})$\;
			
			\textbf{Update} $w=w-\eta\nabla \mathcal{L}_{n}$ \; 
			
		}
		\textbf{Update} $R(t) = 1- \mathtt{min}\big\{ \frac{t}{T_{k}}\tau,\tau\big\}$
	}
	\textbf{Output} $w_{f}$ and $w_{g}$.
\end{algorithm}
After obtaining the small-loss instances set $\hat{D}_{n}$ in mini-batch $n$, we calculate the average loss in these examples for further back propagation.
\begin{align}
	\mathcal{L}_{n} = \frac{1}{|\hat{D}_{n}|}{\sum}_{{\bm{x} \in \hat{D}_{n}}}\ell(\bm{x}).
\end{align}
Put all these together, our algorithm Co-matching is described in Algorithm \ref{alg:1}. It consists of the loss calculation (step 5-7), ``small-loss" selection (step 8-9) and parameter update (step 9-10).


\subsection{Augmentation in Co-matching}
\label{sec:aug}

Co-matching leverages two kinds of augmentations: ``weak" and ``strong". In our experiments, weak augmentation is a standard crop-and-flip augmentation strategy. 
As for ``strong" augmentation, we adopt RandAugment \cite{cubuk2020randaugment}, which is based on AutoAugment \cite{cubuk2019autoaugment}. AutoAugment learns an augmentation strategy based on transformations from the Python Imaging Libraries \footnote{https://www.pythonware.com/products/pil/} using reinforcement learning. Given a collection of transformations (e.g., color inversion, contrast adjustment, translation, etc.), RandAugment randomly select $M$ transformations for each sample in a mini-batch. We set $M=2$ in our experiments and explore more options in Section \ref{sec:exp:augmentation}. As originally proposed, RandAugment uses a single fixed global magnitude that controls the severity of all distortions \cite{cubuk2020randaugment}. Instead of optimizing the hyperparameter magnitude by using grid search, we found that sampling a random magnitude from a pre-defined range at each training step (instead of using a fixed global value) works better for learning with noisy labels. 
\subsection{Comparison to other approaches}
We compare Co-matching with other related approaches in Table \ref{tab:compare}. Specifically, Decoupling applies the ``Disagreement" strategy to select instances while other approaches including Co-matching use ``small-loss" criterion. Co-teaching cross-updates parameters of networks to reduce the accumulated error flow. Combining the ``Disagreement" strategy with Co-teaching, Co-teaching+ achieves excellent performance by keeping two networks diverged. JoCoR updates the networks jointly with ``small-loss" instances selected by using ``agreement" strategy. As for Co-matching, we also select ``small-loss" instances and updates the network jointly. Besides, we keep the two networks diverged by feeding different degrees of augmentation and using augmentation anchoring to limit their divergence to some degree. As we will show in Section \ref{sec:ablation}, these components are crucial to Co-matching's success under high levels of noise.
\begin{table}
	\begin{center}
		\resizebox{0.48\textwidth}{!}{
		\begin{tabular}{c|c|c|c|c|c}
			\hline\hline
			 & Decoupling&Co-teaching&Co-teaching+&JoCoR& Co-matching \\
			\hline
			small loss   &$\times$ &$\surd$ & $\surd$ &$\surd$ & $\surd$ \\
			\hline
			cross update   &$\times$ & $\surd$& $\surd$&$\times$ & $\times$\\
			\hline
			joint update & $\times$ & $\times$ & $\times$& $\surd$&  $\surd$ \\
			\hline
			divergence  & $\surd$&$\times$ & $\surd$ &$\times$&$\surd$\\
			\hline
			augmentation anchoring & $\times$&$\times$ & $\times$&$\times$&$\surd$ \\
			\hline \hline
		\end{tabular}
	}
	\end{center}
	\caption{Comparison of state-of-the-art and related techniques with our approach. In the first column, ``small loss": regarding small-loss samples as ``clean" samples, which is based on the memorization effects of deep neural networks; ``cross update": updating parameters in a cross manner instead of a parallel manner; ``joint update": updating the two networks parameters jointly. ``divergence": keeping two classifiers diverged during the whole training epochs. ``augmentation anchoring": encouraging the predictions of a strongly-augmented image to be close to the predictions from a weakly-augmented version of the same image.} \label{tab:compare}
\end{table}

%% file: exp.tex
\section{Experiments}


\subsection{Experimental settings}
\noindent\textbf{Datasets.} We evaluate the efficacy of Co-matching on two benchmarks with simulated label noise, CIFAR-10 and CIFAR-100 \cite{krizhevsky2009learning}, and one real-world dataset, Clothing1M \cite{xiao2015learning}. Clothing1M consists of 1 million training images collected from online shopping websites with noisy labels generated from surrounding texts. Its noise level is estimated at 38.5\%. CIFAR-10 and CIFAR-100 are initially clean. Following \cite{patrini2017making,reed2014training}, we corrupt these datasets manually by label transition matrix $Q$, where $Q_{ij}=\Pr[\hat{y}=j|y=i]$ given that noisy label $\hat{y}$ is flipped from clean label $y$. The matrix Q has two representative label noise models: (1) Symmetric flipping \cite{van2015learning} is generated by uniformly flipping the label to one of the other class label; (2) Asymmetric flipping \cite{patrini2017making} is a simulation of fine-grained classification with noisy labels in real world, where the labelers are more likely to make mistakes only within very similar classes. More details are described in supplementary materials.





\noindent\textbf{Baselines.} We compare Co-matching (Algorithm \ref{alg:1}) with Decoupling \cite{malach2017decoupling}, Co-teaching \cite{han2018co}, Co-teaching+ \cite{yu2019does} and JoCoR \cite{wei2020combating}, and implement all methods with default parameters by Pytorch. Note that all compared algorithms do not use extra techniques such as mixup, Gaussian mixture model, temperature sharpening and temporal ensembling to improve the performance.


\begin{figure*}[t]
	\begin{center}
		\includegraphics[width=0.9\linewidth]{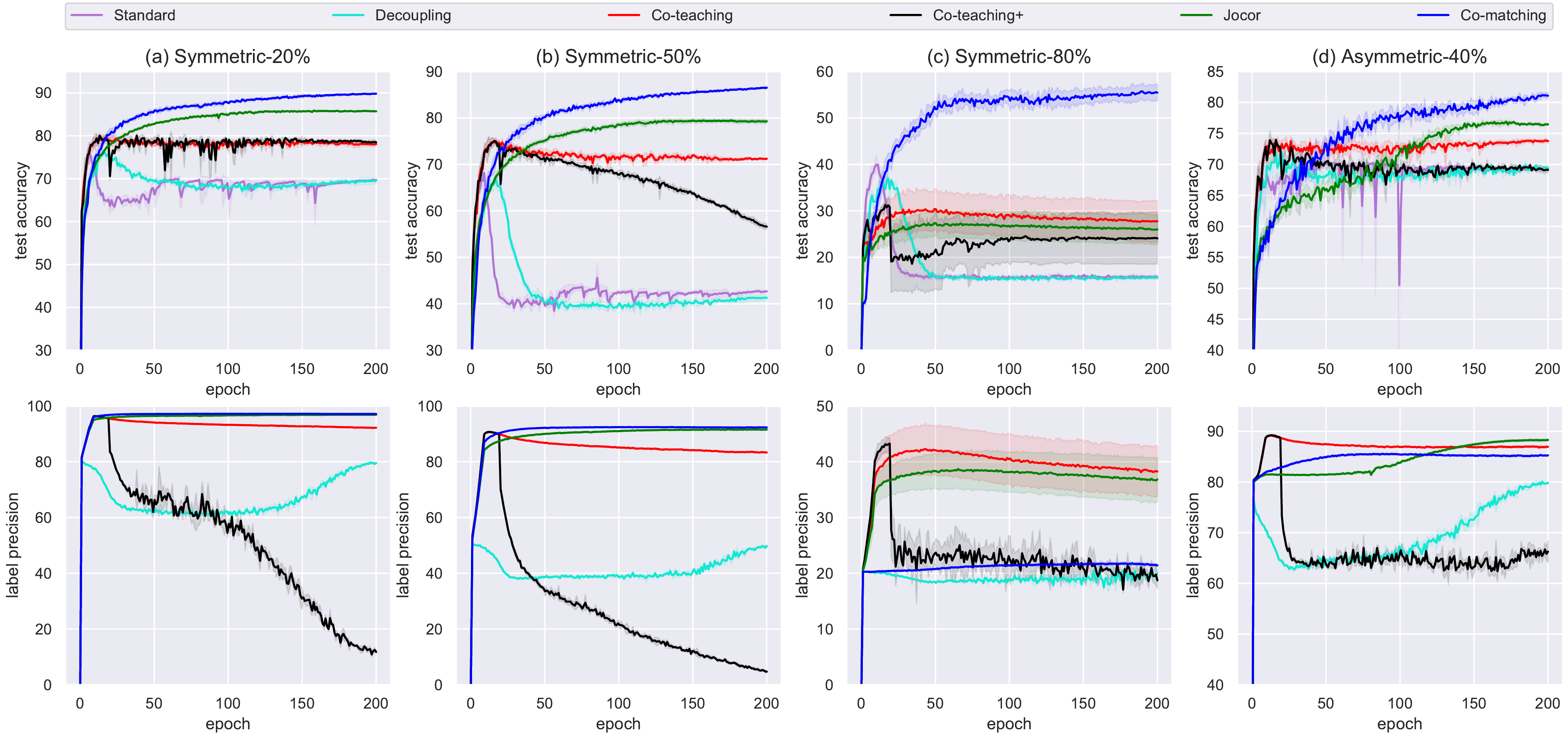}
	\end{center}
	\vspace{-0.5em}
	\caption{Results on CIFAR-10 dataset. Top: test accuracy(\%) vs. epochs; bottom: label precision(\%) vs. epochs.}
	\label{fig:cifar10}
\end{figure*}

\begin{table*}
	\begin{center}
		\resizebox{0.9\textwidth}{!}{
			\begin{tabular}{l|ccccc|c}
				\hline\hline
				Noise ratio & Standard& Decoupling&Co-teaching&Co-teaching+&JoCoR& Co-Matching \\
				\hline
				Symmetric-20\% & 69.57 $\pm$ 0.20  & 69.55 $\pm$ 0.20& 78.07 $\pm$ 0.24& 78.66 $\pm$ 0.20 &85.69 $\pm$ 0.06 & \textbf{89.78} $\pm$ 0.13\\
				\hline
				Symmetric-50\% & 42.48 $\pm$ 0.35 &41.44 $\pm$ 0.46 & 71.54 $\pm$ 0.17 &57.13 $\pm$ 0.46 &79.32 $\pm$ 0.37  &\textbf{86.42} $\pm$ 0.18\\
				\hline
				Symmetric-80\% & 15.79 $\pm$ 0.37&  15.64 $\pm$ 0.42&27.71 $\pm$ 4.39& 24.13 $\pm$ 5.54 & 25.97 $\pm$ 3.11& \textbf{55.42} $\pm$ 1.68\\
				\hline
				Asymmetric-40\% & 69.36 $\pm$ 0.23 & 69.46 $\pm$ 0.08 &73.75 $\pm$ 0.34 & 69.03 $\pm$ 0.30& 76.38 $\pm$ 0.32& \textbf{81.00} $\pm$ 0.46\\
				\hline \hline
			\end{tabular}
		}
	\end{center}
	\vspace{-0.5em}
	\caption{Average test accuracy (\%) on CIFAR-10 over the last 10 epochs. Bold indicates best performance.} \label{table:cifar10}
\end{table*}
\noindent\textbf{Network Structure.} For CIFAR-10 and CIFAR-100, we use a 7-layer network architecture follows \cite{wei2020combating}. The detail information can be found in supplementary material. As for Clothing1M, we use ResNet18 \cite{he2016deep} with ImageNet pretrained weights. 

\noindent\textbf{Optimizer.} For CIFAR-10 and CIFAR-100, Adam optimizer (momentum=0.9) is used with an initial learning rate of 0.001, and the batch size is set to 128. We run 200 epochs in total and linearly decay learning rate to zero from 80 to 200 epochs. For Clothing1M, we also use Adam optimizer (momentum=0.9) and set batch size to 64. We run 20 epochs in total and set learning rate to $8 \times 10^{-4}$, $5 \times 10^{-4}$ and $5 \times 10^{-5}$ for 5, 5 and 10 epochs respectively. 

\noindent\textbf{Initialization.} Following \cite{han2018co,wei2020combating,yu2019does}, we assume the noise rate $\epsilon$ is known. We set ratio of the small-loss samples $R(t)=1-\min\big\{  \frac{t}{T_{k}}\tau,\tau \big\}$, where $T_{k}=10$ and $\tau=\epsilon$ for all datasets. If $\epsilon$ is not known in advance, $\epsilon$ can be inferred using validation sets \cite{liu2015classification,yu2018efficient}. We find that in extremely noisy case (i.e., 80\%), selecting more small-loss instances in Co-matching achieves better generalization performance. For $\lambda$ in our loss function (\ref{eq1}), we search it in [0.05,0.10,$\dots$,0.95] with a clean validation set for best performance. When validation set is also with noisy labels, we apply the small-loss selection to choose a clean subset for validation. More information about hyperparameter sensitivity analysis can be found in supplementary materials.
\begin{figure*}[t]
	\begin{center}
		\includegraphics[width=0.9\linewidth]{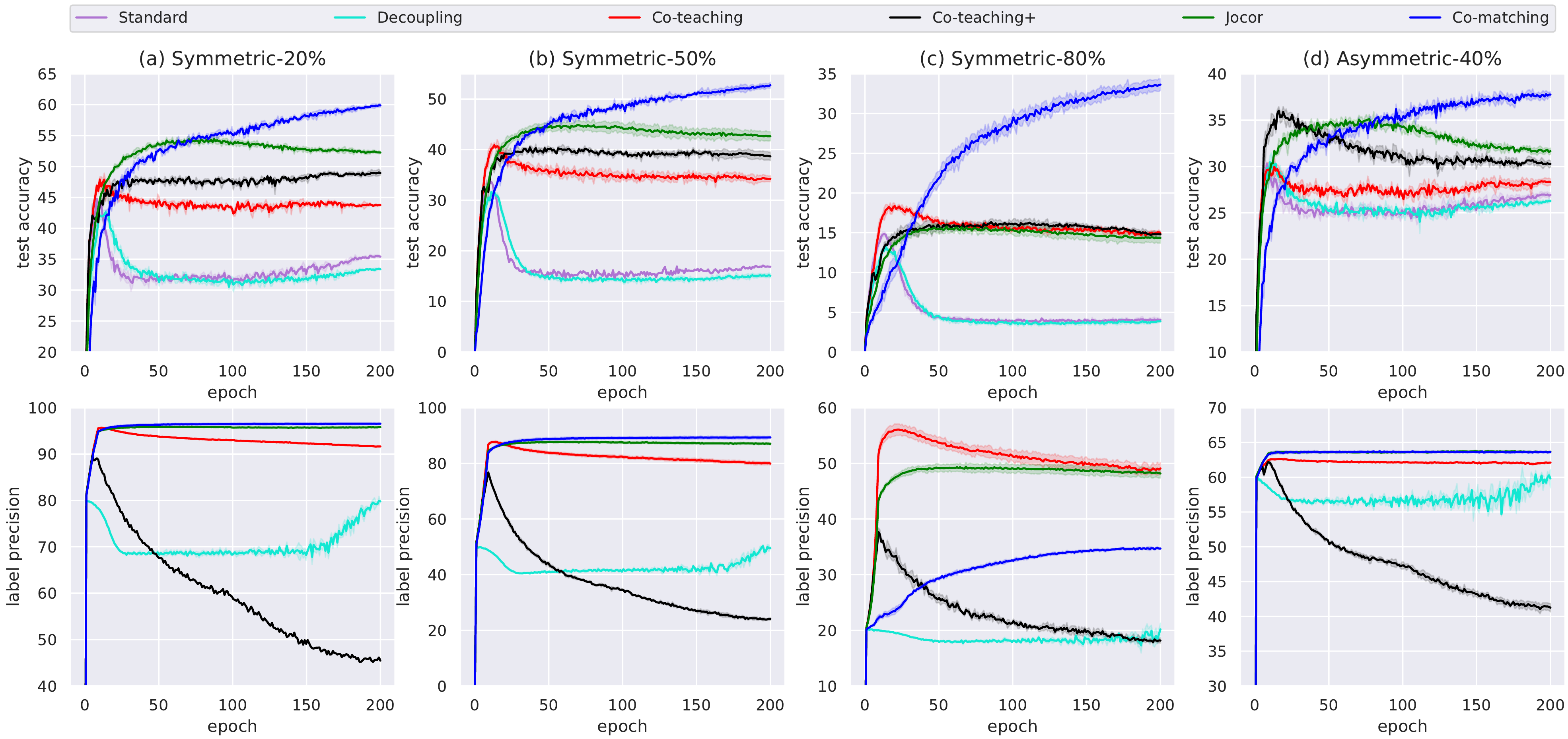}
	\end{center}
	\vspace{-0.5em}
	\caption{Results on CIFAR-100 dataset. Top: test accuracy(\%) vs. epochs; bottom: label precision(\%) vs. epochs.}
	\label{fig:cifar100}
\end{figure*}

\begin{table*}
	\begin{center}
		\resizebox{0.9\textwidth}{!}{
			\begin{tabular}{l|ccccc|c}
				\hline\hline
				Noise ratio & Standard& Decoupling&Co-teaching&Co-teaching+&JoCoR& Co-Matching \\
				\hline
				Symmetric-20\% & 35.46 $\pm$ 0.25 & 33.21 $\pm$ 0.22& 43.71 $\pm$ 0.20&49.15 $\pm$ 0.24 &52.43 $\pm$ 0.20 & \textbf{59.76} $\pm$ 0.27\\
				\hline
				Symmetric-50\% &16.87 $\pm$ 0.13 & 15.03 $\pm$ 0.33& 34.30 $\pm$ 0.39&39.08 $\pm$ 0.73  &42.73 $\pm$ 0.96  &\textbf{52.52} $\pm$ 0.48\\
				\hline
				Symmetric-80\% &4.08 $\pm$ 0.21&3.80 $\pm$ 0.01&14.95 $\pm$ 0.15& 15.00 $\pm$ 0.42& 14.41 $\pm$ 0.60& \textbf{33.50} $\pm$ 0.74\\
				\hline
				Asymmetric-40\% & 27.23 $\pm$ 0.45& 26.25 $\pm$ 0.27&28.27 $\pm$ 0.22 &30.45 $\pm$ 0.15 & 31.52 $\pm$ 0.31& \textbf{37.67} $\pm$ 0.35 \\
				\hline \hline
			\end{tabular}
		}
	\end{center}
	\vspace{-0.5em}
	\caption{Average test accuracy (\%) on CIFAR-100 over the last 10 epochs. Bold indicates best performance.} \label{table:cifar100}
\end{table*}
\noindent\textbf{Metrics.} To measure the performance, we use the test accuracy, i.e., \emph{test accuracy} = (\# \emph{of correct predictions}) / (\# \emph{of test dataset}). Higher test accuracy means that the algorithm is more robust to the label noise. Following the \cite{han2018co,wei2020combating}, we also calculate the label precision in each mini-batch, i.e., \emph{label precision} = (\# \emph{of clean labels}) / (\# \emph{of all selected labels}). Specifically, we sample $R(t)$ of small-loss instances in each mini-batch, and then calculate the ratio of clean labels in the small-loss instances. Intuitively, higher label precision means less noisy instances in the mini-batch after sample selection, and the algorithm with higher label precision is also more robust to the label noise \cite{han2018co,wei2020combating}. However, we find that the higher label precision is not necessarily lead to higher test accuracy with extreme label noise, we will explain this phenomenon in Section \ref{sec:exp:comparison}. All experiments are repeated five times. The error bar for STD in each figure has been highlighted as shade.

\subsection{Comparison with the State-of-the-Arts}
\label{sec:exp:comparison}
\noindent\textbf{Results on CIFAR-10.} Figure \ref{fig:cifar10} shows the test accuracy vs. epochs on CIFAR-10 (top) and the label precision vs. epochs (bottom). With different levels of symmetric and asymmetric label noise, we can clearly see the memorization effect of networks. i.e., test accuracy of Standard first reaches a very high level and then gradually decreases due to the networks overfit to noisy labels. Thus, a robust training approach should alleviate or even stop the decreasing trend in test accuracy. On this point, Co-matching consistently outperforms state-of-the-art methods by a large margin across all noise ratios.

We report the test accuracy of different algorithms in detail in Table \ref{table:cifar10}. In the easiest Symmetric-20\% case, all existing approaches except Decoupling perform much better than Standard, which demonstrates their robustness. Compared to the best baseline method JoCoR, Co-matching still achieve more than 4\% improvement. When it goes to Symmetric-50\% case and Asymmetric 40\% case, Co-teaching+ and Decoupling begin to fail a lot while other methods still work fine, especially Co-matching and JoCoR. However, similar to Co-teaching, JoCoR cannot combat with the hardest Symmetric-80\% case, where it only achieves 25.97\%, which is even worse than Co-teaching (27.71\%). In short, JoCoR reduces to Co-teaching in function and suffers the same problem which the two classifiers converge to consensus. We hypothesize the reason is that under high-levels of label noise, there is not enough supervision from clean labels, which leads to the networks making the same wrong predictions even with low confidence. However, Co-matching achieves the best average classification accuracy (55.42\%) again. 

We also plot label precision vs. epochs at the bottom of Figure \ref{fig:cifar10}. Only Decoupling, Co-teaching, Co-teaching+, JoCoR and Co-matching are considered here, as they include sample selection during training. First, we can see Co-matching, JoCoR and Co-teaching can successfully pick clean instances out in Symmetric-20 \%, Symmetric-50\% and Asymmetric-40\% cases. Note that Co-matching not only reaches high label precision in these three cases but also performs better and better with the increase of epochs. Decoupling and Co-teaching+ fail in selecting clean samples, because ``Disagreement" strategy does not guarantee to select clean samples, as mentioned in Section \ref{sec:relate}. However, an interesting phenomenon is that high label precision does not necessarily lead to high test accuracy especially under high-levels of label noise. For example, in Symmetric-80\% case, the label precision of Co-matching is much lower than Co-teaching and JoCoR, while the test accuracy is much higher than Co-teaching and JoCoR. More similar cases can be found in results on CIFAR-100.




\noindent\textbf{Results on CIFAR-100.}
The test accuracy and label precision vs. epochs are shown in Figure \ref{fig:cifar100}. The test accuracy is shown in Table \ref{table:cifar100}. Similarly, Co-matching still achieves highest test accuracy in all noise cases. In the hardest Symmetric-80\% case, Co-teaching, Co-teaching+ and JoCoR tie together, while Co-matching gets much higher test accuracy. When it turns to Asymmetric-40\% case, Co-teaching+ and JoCoR achieve the better performance first, while Co-matching gradually surpasses these methods after 75 epochs. Overall, the result shows that Co-matching has better generalization ability than state-of-the-art approaches. 

As we discussed before, high label precision does not necessarily lead to high test accuracy. The results in Figure \ref{fig:cifar100} also verify it. As we can see in all four noise rate cases, Co-teaching has much higher label precision than Co-teaching+, while the test accuracy of Co-teaching is lower than Co-teaching+. In the training process, compared to repetitive clean samples, the samples which are near the margins with low confidence and relatively high loss may contribute more towards improving the model's generalization ability. This explains why Co-matching has low label precision but higher test accuracy in Symmetric-80\% case.


\noindent\textbf{Results on Clothing1M.} As shown in Table \ref{table:clothing1M}, \emph{best} denotes the epoch where the validation accuracy is optimal, and \emph{last} denotes the test accuracy at the end of training. Co-matching outperforms the state-of-the-art methods by a large margin on both \emph{best} and \emph{last}, e.g., improving the accuracy from 66.95\% to 70.71\% over Standard, better than best baseline method by 0.92\%.


\subsection{Composition of data augmentation}
\label{sec:exp:augmentation}

We study the impact of data augmentation systematically by considering several common augmentations. One type of augmentation involves spatial/geometric transformation, such as cropping, flipping, rotation and cutout. The other type of augmentation involves appearance transformation, such as color distortion (e.g. brightness and contrast) and Gaussian blur. Since Clothing1M images are of different sizes, we always use cropping as a base transformation. We explore various ``weak" augmentation by combining cropping with other augmentations. As for ``strong" augmentation, we use RandAugment \cite{cubuk2020randaugment}, which randomly select $M$ transformations from a set $\mathcal{S}$ for each sample in a mini-batch. We denote RandAugment as $\mathcal{S}(M)$. In our experiment, $\mathcal{S}=\{$\emph{Contrast}, \emph{Equalize}, \emph{Invert}, \emph{Rotate}, \emph{Posterize}, \emph{Solarize}, \emph{Color}, \emph{Brightness}, \emph{Sharpness}, \emph{ShearX}, \emph{ShearY}, \emph{Cutout}, \emph{TranslateX}, \emph{TranslateY}, \emph{Gaussian Blur}$\}$. 

\begin{figure}[t]
	\begin{center}
		\includegraphics[width=1\linewidth]{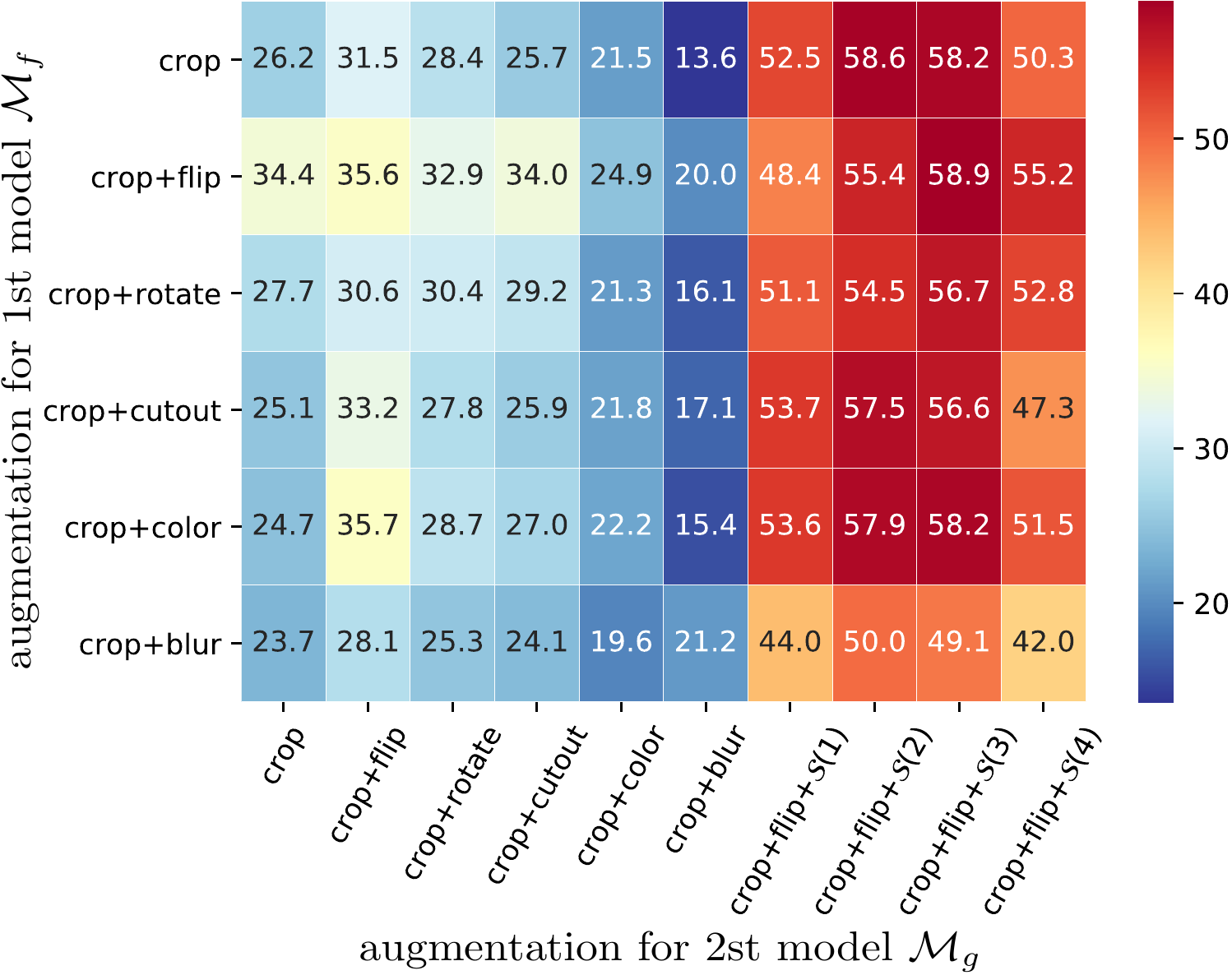}
	\end{center}
	\caption{Average test accuracy(\%) over various combinations of augmentations on CIFAR-10 with Symmetric-80\% label noise.}
	\label{fig:aug}
\end{figure}

Figure \ref{fig:aug} shows the results under composition of transformations. We observe that using 
``weak" augmentation for both models does not work much better than simple cropping. However, the performance of Co-matching benefits a lot by using stronger augmentations (e.g. add $\mathcal{S}(2)$ and $\mathcal{S}(3)$) on second model. We conclude that in extreme label noise case, the unsupervised matching loss requires to use stronger augmentation on model $\mathcal{M}_{g}$ to improve its effect.



\subsection{Ablation Study}
\label{sec:ablation}
In this section, we perform an ablation study for analyzing the effect of using two networks and augmentation anchoring in Co-matching. The experiment is conducted on CIFAR-10 with two cases: Symmetric-50\% and Symmetric-80\%. To verify the effect of using two classifiers, we introduce Standard enhanced by ``small-loss" selection (abbreviated as Standard+), Co-teaching and JoCoR to join the comparison. Recall that JoCoR selects instances by the joint loss while Co-teaching uses cross-update approach to reduce the accumulated error \cite{han2018co,wei2020combating}. Besides, we simply set $\lambda=0$ in (\ref{eq1}) to see the influence of removing augmentation anchoring (abbreviated as Co-matching-). 

\begin{figure}[t]
	\begin{center}
		\includegraphics[width=1.0\linewidth]{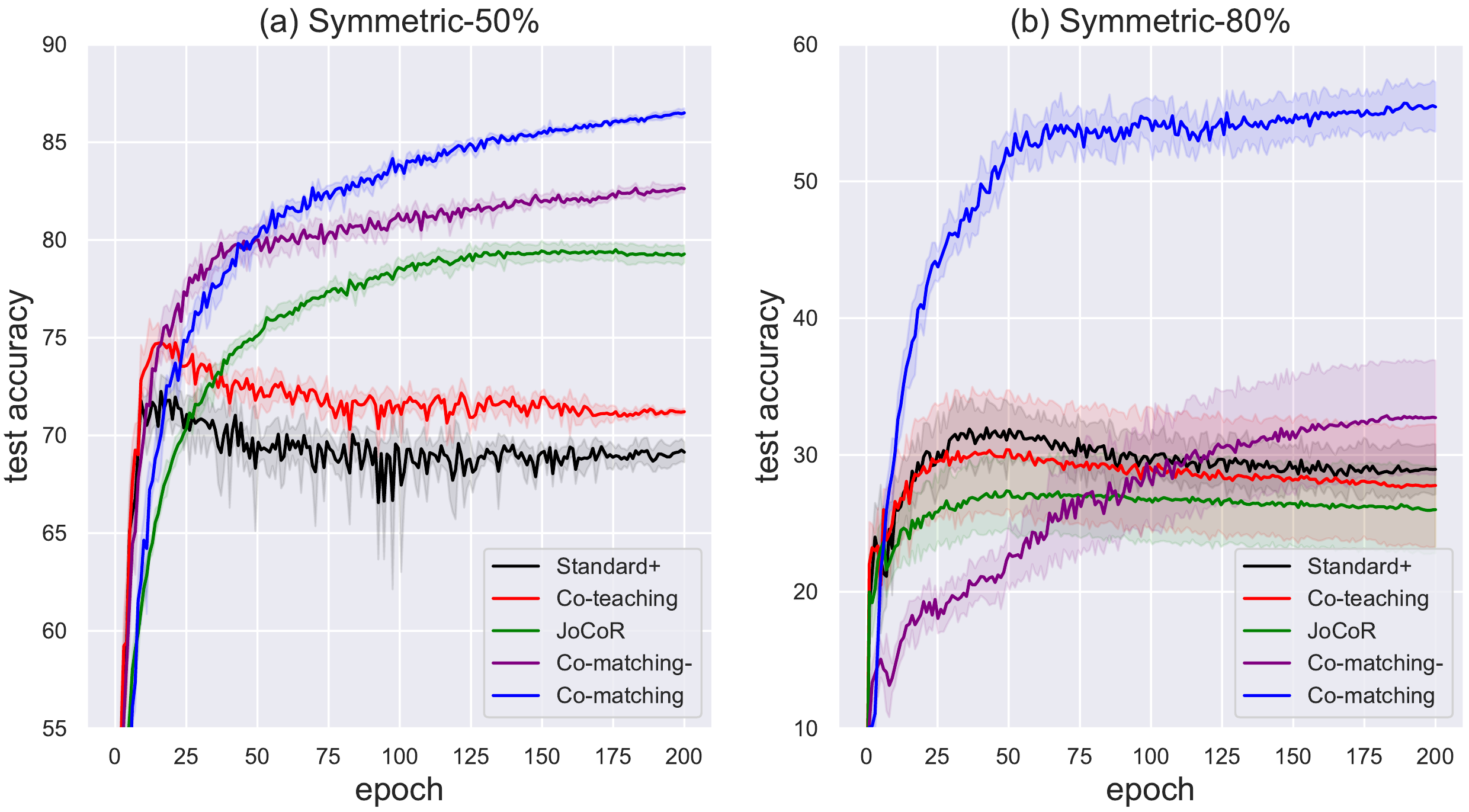}
	\end{center}
	\caption{Results of ablation study on CIFAR-10}
	\label{fig:ablation}
\end{figure}

\begin{table}
	\begin{center}
		\resizebox{0.46\textwidth}{!}{
			\begin{tabular}{p{3cm}<{\centering}|p{2cm}<{\centering}|p{2cm}<{\centering}}
				\hline\hline
				Methods & \emph{best}& \emph{last} \\
				\hline
				Standard & 67.74 & 66.95  \\
				Decoupling & 67.71&66.78  \\
				Co-teaching &69.05&68.99 \\
				Co-teaching+ & 67.84& 67.68 \\
				JoCoR & 70.30& 69.79 \\
				\hline
				Co-matching & \textbf{71.03} & \textbf{70.71} \\
				\hline \hline
			\end{tabular}
		}
	\end{center}
	\caption{Test accuracy (\%) on Clothing1M with ResNet18. Bold indicates best performance.} \label{table:clothing1M}
\end{table}
The results of their test accuracy vs. epochs are shown in Figure \ref{fig:ablation}. In Symmetric-50\% case, both Co-teaching and Standard+ keep a downward tendency after increasing to the highest point, which indicates they are still prone to memorizing noisy labels even with ``small-loss" update. Besides, it proves the effect of using two networks as Co-teaching performs better than Standard+. JoCoR consistently outperforms Co-teaching, which verifies the conclusion in \cite{wei2020combating} that joint-update is more efficient than cross-update in Co-teaching. However, things start to change in Symmetric-80\% case. Co-teaching and Standard+ remain the same trend as these for Symmetric-50\% case, but JoCoR performs even worse than Co-teaching and Standard+. Since two networks in JoCoR would be more and more similar due to the effect of Co-regularization, using joint-update works the same as cross-update. 

In both cases, Co-matching consistently outperforms Co-matching- and other methods, which validates that augmentation anchoring can strongly prevent neural networks from memorizing noisy labels as it reduces the dependency on noisy labels. In addition, using weak and strong augmentations alone may not promise to improve test accuracy. As we can see in Symmetric-80\% case, Co-matching- only slightly outperforms Standard+ after 115 epochs, which also demonstrates that only using the classification loss in (\ref{eq2}) doesn't provide a strong supervision since the most noisy labels are incorrect in this case. 







%% file: iccv_co_matching.bbl
\begin{thebibliography}{10}\itemsep=-1pt

\bibitem{arazo2019unsupervised}
Eric Arazo, Diego Ortego, Paul Albert, Noel O’Connor, and Kevin Mcguinness.
\newblock Unsupervised label noise modeling and loss correction.
\newblock In {\em International Conference on Machine Learning}, pages
  312--321, 2019.

\bibitem{arpit2017closer}
Devansh Arpit, Stanislaw~K Jastrzebski, Nicolas Ballas, David Krueger, Emmanuel
  Bengio, Maxinder~S Kanwal, Tegan Maharaj, Asja Fischer, Aaron~C Courville,
  Yoshua Bengio, et~al.
\newblock A closer look at memorization in deep networks.
\newblock In {\em ICML}, 2017.

\bibitem{bachman2014learning}
Philip Bachman, Ouais Alsharif, and Doina Precup.
\newblock Learning with pseudo-ensembles.
\newblock In {\em Advances in neural information processing systems}, pages
  3365--3373, 2014.

\bibitem{bengio2009curriculum}
Yoshua Bengio, J{\'e}r{\^o}me Louradour, Ronan Collobert, and Jason Weston.
\newblock Curriculum learning.
\newblock In {\em Proceedings of the 26th annual international conference on
  machine learning}, pages 41--48, 2009.

\bibitem{berthelot2019remixmatch}
David Berthelot, Nicholas Carlini, Ekin~D Cubuk, Alex Kurakin, Kihyuk Sohn, Han
  Zhang, and Colin Raffel.
\newblock Remixmatch: Semi-supervised learning with distribution alignment and
  augmentation anchoring.
\newblock {\em arXiv preprint arXiv:1911.09785}, 2019.

\bibitem{blum1998combining}
Avrim Blum and Tom Mitchell.
\newblock Combining labeled and unlabeled data with co-training.
\newblock In {\em Proceedings of the eleventh annual conference on
  Computational learning theory}, pages 92--100, 1998.

\bibitem{chen2019understanding}
Pengfei Chen, Ben~Ben Liao, Guangyong Chen, and Shengyu Zhang.
\newblock Understanding and utilizing deep neural networks trained with noisy
  labels.
\newblock In {\em International Conference on Machine Learning}, pages
  1062--1070, 2019.

\bibitem{cubuk2019autoaugment}
Ekin~D Cubuk, Barret Zoph, Dandelion Mane, Vijay Vasudevan, and Quoc~V Le.
\newblock Autoaugment: Learning augmentation strategies from data.
\newblock In {\em Proceedings of the IEEE conference on computer vision and
  pattern recognition}, pages 113--123, 2019.

\bibitem{cubuk2020randaugment}
Ekin~D Cubuk, Barret Zoph, Jonathon Shlens, and Quoc~V Le.
\newblock Randaugment: Practical automated data augmentation with a reduced
  search space.
\newblock In {\em Proceedings of the IEEE/CVF Conference on Computer Vision and
  Pattern Recognition Workshops}, pages 702--703, 2020.

\bibitem{ding2018semi}
Yifan Ding, Liqiang Wang, Deliang Fan, and Boqing Gong.
\newblock A semi-supervised two-stage approach to learning from noisy labels.
\newblock In {\em 2018 IEEE Winter Conference on Applications of Computer
  Vision (WACV)}, pages 1215--1224. IEEE, 2018.

\bibitem{ghosh2017robust}
Aritra Ghosh, Himanshu Kumar, and PS Sastry.
\newblock Robust loss functions under label noise for deep neural networks.
\newblock In {\em Proceedings of the Thirty-First AAAI Conference on Artificial
  Intelligence}, pages 1919--1925, 2017.

\bibitem{goldberger2016training}
Jacob Goldberger and Ehud Ben-Reuven.
\newblock Training deep neural-networks using a noise adaptation layer.
\newblock 2016.

\bibitem{grandvalet2005semi}
Yves Grandvalet and Yoshua Bengio.
\newblock Semi-supervised learning by entropy minimization.
\newblock In {\em Advances in neural information processing systems}, pages
  529--536, 2005.

\bibitem{han2018co}
Bo Han, Quanming Yao, Xingrui Yu, Gang Niu, Miao Xu, Weihua Hu, Ivor Tsang, and
  Masashi Sugiyama.
\newblock Co-teaching: Robust training of deep neural networks with extremely
  noisy labels.
\newblock In {\em Advances in neural information processing systems}, pages
  8527--8537, 2018.

\bibitem{he2016deep}
Kaiming He, Xiangyu Zhang, Shaoqing Ren, and Jian Sun.
\newblock Deep residual learning for image recognition.
\newblock In {\em Proceedings of the IEEE conference on computer vision and
  pattern recognition}, pages 770--778, 2016.

\bibitem{hendrycks2018using}
Dan Hendrycks, Mantas Mazeika, Duncan Wilson, and Kevin Gimpel.
\newblock Using trusted data to train deep networks on labels corrupted by
  severe noise.
\newblock In {\em Advances in neural information processing systems}, pages
  10456--10465, 2018.

\bibitem{jiang2018mentornet}
Lu Jiang, Zhengyuan Zhou, Thomas Leung, Li-Jia Li, and Li Fei-Fei.
\newblock Mentornet: Learning data-driven curriculum for very deep neural
  networks on corrupted labels.
\newblock In {\em International Conference on Machine Learning}, pages
  2304--2313, 2018.

\bibitem{krizhevsky2009learning}
Alex Krizhevsky, Geoffrey Hinton, et~al.
\newblock Learning multiple layers of features from tiny images.
\newblock 2009.

\bibitem{krizhevsky2012imagenet}
Alex Krizhevsky, Ilya Sutskever, and Geoffrey~E Hinton.
\newblock Imagenet classification with deep convolutional neural networks.
\newblock In {\em Advances in neural information processing systems}, pages
  1097--1105, 2012.

\bibitem{laine2016temporal}
Samuli Laine and Timo Aila.
\newblock Temporal ensembling for semi-supervised learning.
\newblock {\em arXiv preprint arXiv:1610.02242}, 2016.

\bibitem{li2019dividemix}
Junnan Li, Richard Socher, and Steven~CH Hoi.
\newblock Dividemix: Learning with noisy labels as semi-supervised learning.
\newblock In {\em International Conference on Learning Representations}, 2020.

\bibitem{li2017webvision}
Wen Li, Limin Wang, Wei Li, Eirikur Agustsson, and Luc Van~Gool.
\newblock Webvision database: Visual learning and understanding from web data.
\newblock {\em arXiv preprint arXiv:1708.02862}, 2017.

\bibitem{liu2015classification}
Tongliang Liu and Dacheng Tao.
\newblock Classification with noisy labels by importance reweighting.
\newblock {\em IEEE Transactions on pattern analysis and machine intelligence},
  38(3):447--461, 2015.

\bibitem{liu2019improving}
Xihui Liu, Zihao Wang, Jing Shao, Xiaogang Wang, and Hongsheng Li.
\newblock Improving referring expression grounding with cross-modal
  attention-guided erasing.
\newblock In {\em Proceedings of the IEEE Conference on Computer Vision and
  Pattern Recognition}, pages 1950--1959, 2019.

\bibitem{lyu2019curriculum}
Yueming Lyu and Ivor~W Tsang.
\newblock Curriculum loss: Robust learning and generalization against label
  corruption.
\newblock In {\em International Conference on Learning Representations}, 2019.

\bibitem{ma2018dimensionality}
Xingjun Ma, Yisen Wang, Michael~E Houle, Shuo Zhou, Sarah Erfani, Shutao Xia,
  Sudanthi Wijewickrema, and James Bailey.
\newblock Dimensionality-driven learning with noisy labels.
\newblock In {\em International Conference on Machine Learning}, pages
  3355--3364, 2018.

\bibitem{malach2017decoupling}
Eran Malach and Shai Shalev-Shwartz.
\newblock Decoupling" when to update" from" how to update".
\newblock In {\em Advances in Neural Information Processing Systems}, pages
  960--970, 2017.

\bibitem{mclachlan1975iterative}
Geoffrey~J McLachlan.
\newblock Iterative reclassification procedure for constructing an
  asymptotically optimal rule of allocation in discriminant analysis.
\newblock {\em Journal of the American Statistical Association},
  70(350):365--369, 1975.

\bibitem{nguyen2020self}
Tam Nguyen, C Mummadi, T Ngo, L Beggel, and Thomas Brox.
\newblock Self: learning to filter noisy labels with self-ensembling.
\newblock In {\em International Conference on Learning Representations (ICLR)},
  2020.

\bibitem{patrini2017making}
Giorgio Patrini, Alessandro Rozza, Aditya Krishna~Menon, Richard Nock, and
  Lizhen Qu.
\newblock Making deep neural networks robust to label noise: A loss correction
  approach.
\newblock In {\em Proceedings of the IEEE Conference on Computer Vision and
  Pattern Recognition}, pages 1944--1952, 2017.

\bibitem{reed2014training}
Scott Reed, Honglak Lee, Dragomir Anguelov, Christian Szegedy, Dumitru Erhan,
  and Andrew Rabinovich.
\newblock Training deep neural networks on noisy labels with bootstrapping.
\newblock {\em arXiv preprint arXiv:1412.6596}, 2014.

\bibitem{ren2018learning}
Mengye Ren, Wenyuan Zeng, Bin Yang, and Raquel Urtasun.
\newblock Learning to reweight examples for robust deep learning.
\newblock In {\em International Conference on Machine Learning}, pages
  4334--4343, 2018.

\bibitem{sajjadi2016regularization}
Mehdi Sajjadi, Mehran Javanmardi, and Tolga Tasdizen.
\newblock Regularization with stochastic transformations and perturbations for
  deep semi-supervised learning.
\newblock In {\em Advances in neural information processing systems}, pages
  1163--1171, 2016.

\bibitem{scudder1965probability}
H Scudder.
\newblock Probability of error of some adaptive pattern-recognition machines.
\newblock {\em IEEE Transactions on Information Theory}, 11(3):363--371, 1965.

\bibitem{sohn2020fixmatch}
Kihyuk Sohn, David Berthelot, Chun-Liang Li, Zizhao Zhang, Nicholas Carlini,
  Ekin~D Cubuk, Alex Kurakin, Han Zhang, and Colin Raffel.
\newblock Fixmatch: Simplifying semi-supervised learning with consistency and
  confidence.
\newblock {\em arXiv preprint arXiv:2001.07685}, 2020.

\bibitem{song2019selfie}
Hwanjun Song, Minseok Kim, and Jae-Gil Lee.
\newblock Selfie: Refurbishing unclean samples for robust deep learning.
\newblock In {\em International Conference on Machine Learning}, pages
  5907--5915, 2019.

\bibitem{sukhbaatar2014learning}
Sainbayar Sukhbaatar and Rob Fergus.
\newblock Learning from noisy labels with deep neural networks.
\newblock {\em arXiv preprint arXiv:1406.2080}, 2(3):4, 2014.

\bibitem{szegedy2015going}
Christian Szegedy, Wei Liu, Yangqing Jia, Pierre Sermanet, Scott Reed, Dragomir
  Anguelov, Dumitru Erhan, Vincent Vanhoucke, and Andrew Rabinovich.
\newblock Going deeper with convolutions.
\newblock In {\em Proceedings of the IEEE conference on computer vision and
  pattern recognition}, pages 1--9, 2015.

\bibitem{tanaka2018joint}
Daiki Tanaka, Daiki Ikami, Toshihiko Yamasaki, and Kiyoharu Aizawa.
\newblock Joint optimization framework for learning with noisy labels.
\newblock In {\em Proceedings of the IEEE Conference on Computer Vision and
  Pattern Recognition}, pages 5552--5560, 2018.

\bibitem{van2015learning}
Brendan Van~Rooyen, Aditya Menon, and Robert~C Williamson.
\newblock Learning with symmetric label noise: The importance of being
  unhinged.
\newblock In {\em Advances in Neural Information Processing Systems}, pages
  10--18, 2015.

\bibitem{wang2018iterative}
Yisen Wang, Weiyang Liu, Xingjun Ma, James Bailey, Hongyuan Zha, Le Song, and
  Shu-Tao Xia.
\newblock Iterative learning with open-set noisy labels.
\newblock In {\em Proceedings of the IEEE Conference on Computer Vision and
  Pattern Recognition}, pages 8688--8696, 2018.

\bibitem{wang2019symmetric}
Yisen Wang, Xingjun Ma, Zaiyi Chen, Yuan Luo, Jinfeng Yi, and James Bailey.
\newblock Symmetric cross entropy for robust learning with noisy labels.
\newblock In {\em Proceedings of the IEEE International Conference on Computer
  Vision}, pages 322--330, 2019.

\bibitem{wei2020combating}
Hongxin Wei, Lei Feng, Xiangyu Chen, and Bo An.
\newblock Combating noisy labels by agreement: A joint training method with
  co-regularization.
\newblock In {\em Proceedings of the IEEE/CVF Conference on Computer Vision and
  Pattern Recognition}, pages 13726--13735, 2020.

\bibitem{xia2019anchor}
Xiaobo Xia, Tongliang Liu, Nannan Wang, Bo Han, Chen Gong, Gang Niu, and
  Masashi Sugiyama.
\newblock Are anchor points really indispensable in label-noise learning?
\newblock In {\em Advances in Neural Information Processing Systems}, pages
  6838--6849, 2019.

\bibitem{xiao2015learning}
Tong Xiao, Tian Xia, Yi Yang, Chang Huang, and Xiaogang Wang.
\newblock Learning from massive noisy labeled data for image classification.
\newblock In {\em Proceedings of the IEEE conference on computer vision and
  pattern recognition}, pages 2691--2699, 2015.

\bibitem{yi2019probabilistic}
Kun Yi and Jianxin Wu.
\newblock Probabilistic end-to-end noise correction for learning with noisy
  labels.
\newblock In {\em Proceedings of the IEEE Conference on Computer Vision and
  Pattern Recognition}, pages 7017--7025, 2019.

\bibitem{yu2019does}
X Yu, B Han, J Yao, G Niu, IW Tsang, and M Sugiyama.
\newblock How does disagreement help generalization against label corruption?
\newblock In {\em 36th International Conference on Machine Learning, ICML
  2019}, 2019.

\bibitem{yu2018efficient}
Xiyu Yu, Tongliang Liu, Mingming Gong, Kayhan Batmanghelich, and Dacheng Tao.
\newblock An efficient and provable approach for mixture proportion estimation
  using linear independence assumption.
\newblock In {\em Proceedings of the IEEE Conference on Computer Vision and
  Pattern Recognition}, pages 4480--4489, 2018.

\bibitem{yu2018learning}
Xiyu Yu, Tongliang Liu, Mingming Gong, and Dacheng Tao.
\newblock Learning with biased complementary labels.
\newblock In {\em Proceedings of the European Conference on Computer Vision
  (ECCV)}, pages 68--83, 2018.

\bibitem{zhang2016understanding}
Chiyuan Zhang, Samy Bengio, Moritz Hardt, Benjamin Recht, and Oriol Vinyals.
\newblock Understanding deep learning requires rethinking generalization.
\newblock {\em arXiv preprint arXiv:1611.03530}, 2016.

\bibitem{zhang2018generalized}
Zhilu Zhang and Mert Sabuncu.
\newblock Generalized cross entropy loss for training deep neural networks with
  noisy labels.
\newblock In {\em Advances in neural information processing systems}, pages
  8778--8788, 2018.

\end{thebibliography}
